\documentclass{article}

\PassOptionsToPackage{numbers,compress}{natbib}
\usepackage[preprint]{neurips_2026}
\usepackage{booktabs}
\usepackage{makecell}
\usepackage[table]{xcolor}
\usepackage{tabularx}
\usepackage{array}

\usepackage[utf8]{inputenc} 
\usepackage[T1]{fontenc}    
\usepackage{hyperref}       
\usepackage{url}            
\usepackage{booktabs}       
\usepackage{amsfonts}       
\usepackage{nicefrac}       
\usepackage{microtype}      
\usepackage{xcolor}         
\usepackage{booktabs}
\usepackage{multirow}
\usepackage{xcolor}
\usepackage{graphicx}
\usepackage{amsmath} 
\usepackage{colortbl}
\title{ChainSpace: A Chained-Reasoning Paradigm for \\ Spatial Intelligence}

\author{%
Xiaohan Zhang\textsuperscript{1,2,*},
Feng Gu\textsuperscript{2,*,\textdagger},
Xudong Rao\textsuperscript{2},
Xuhao Pan\textsuperscript{2},
Tao Wei\textsuperscript{2},
Zhou Pan\textsuperscript{2, \textdagger},
Kun Zhan\textsuperscript{2}
\\[2mm]
\textsuperscript{1}College of Information Science and Electronic Engineering, Zhejiang University\\
\textsuperscript{2}Li Auto Inc.\\
\\[-2mm]
\textsuperscript{*}Equal contribution. 
\textsuperscript{\textdagger}Corresponding authors.\\
\texttt{zhangxh2023@zju.edu.cn, gufeng@lixiang.com}
}

\begin{document}

\maketitle

\begin{abstract}
Spatial intelligence requires foundation models to maintain coherent spatial state across interactions with the physical world.
However, existing data-centric approaches typically treat spatial reasoning as independent question-answer instances, enabling shortcut-based answering and providing limited supervision for persistent spatial understanding.
To address this, we introduce \texttt{ChainSpace}, a chained-reasoning paradigm that structures spatial reasoning as a state-preserving multi-round process.
In this paradigm, spatial questions are organized into logically constrained and jointly consistent chains, where later questions depend on spatial constraints established in earlier rounds.
Following this principle, we instantiate \texttt{ChainSpace-Bench}, a manually annotated real-world multi-round benchmark with a \emph{Chain-Aware Metric}, and \texttt{ChainSpace-Pipeline}, a simulator-based chain-structured supervision generation framework for spatial intelligence training.
Experiments show that \texttt{ChainSpace-Bench} exposes chain-level failures that are not captured by isolated question accuracy.
Additionally, with a relatively small amount of simulator-generated chained data, models trained by \texttt{ChainSpace-Pipeline} achieve the best performance among open-source models on \texttt{ChainSpace-Bench} and transfer competitively to multiple external spatial intelligence benchmarks.
These results establish \texttt{ChainSpace} as an effective paradigm for more faithful evaluation and more data-efficient learning of spatial intelligence.
\end{abstract}

\section{Introduction}
\label{Introduction}
Spatial intelligence is foundational for multimodal large language models~\cite{qwen3, internvl, video-llava} to interact with the physical world, with broad applications in vision-and-language navigation~\cite{Bevbert, LHVLN}, autonomous driving~\cite{RAG-Driver, safeauto}, and robotics~\cite{world-vla, vla-r1}. However, despite rapid progress in multimodal foundation models~\cite{qwen3, internvl, video-llava}, spatial intelligence remains a major challenge~\cite{SenseNova}.

Existing works on spatial intelligence can be roughly grouped into architectural solutions~\cite{sdvlm, spatialrgpt} and data-centric approaches~\cite{vsi-bench, SenseNova, robobrain25}. Among them, data-centric approaches are particularly attractive for general-purpose foundation models, as they evaluate and improve spatial intelligence without altering the underlying model architecture. These approaches span evaluation benchmarks~\cite{vsi-bench, embspatial, MMSI-Bench, viewspatialbench}, which diagnose spatial capabilities through curated test instances, and training-data construction~\cite{SenseNova, robobrain2.0, cambrians}, which improves spatial capability by scaling spatial supervision.

\begin{figure}[t]
    \centering
    \includegraphics[width=0.9\textwidth]{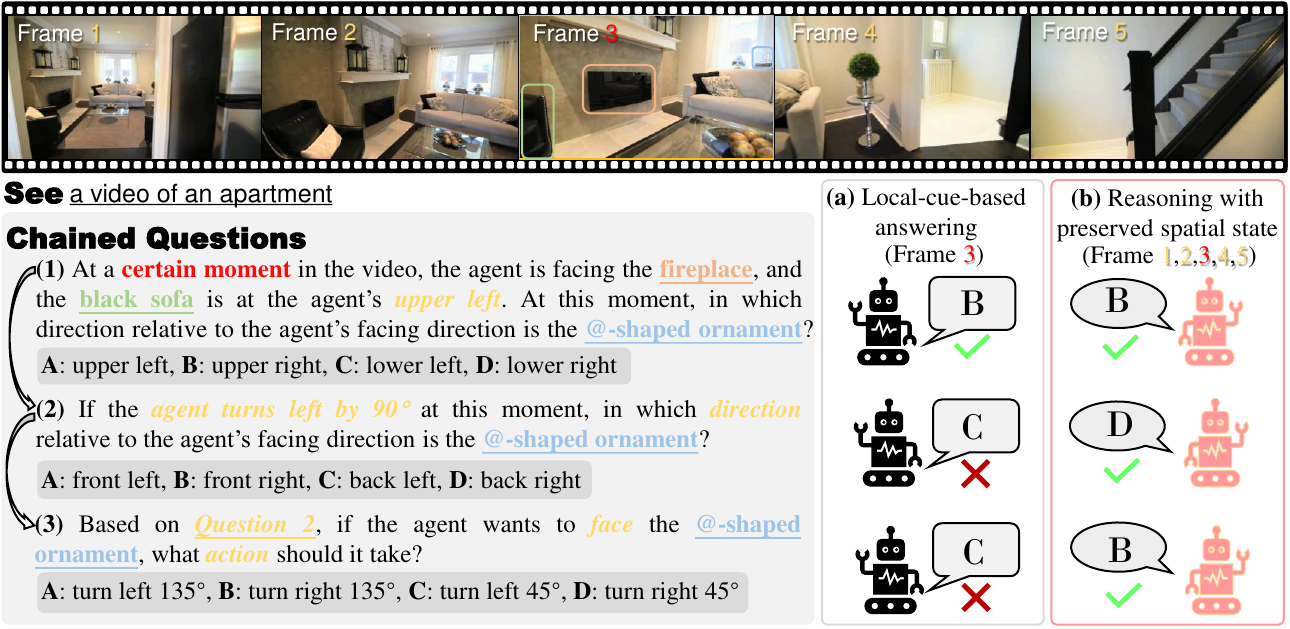}
    \caption{
    Illustration of the \texttt{ChainSpace} paradigm. 
    Given a video trajectory, spatial questions are organized into a logically constrained chain, where later questions depend on spatial constraints established in earlier rounds. 
    (a) A model following an independent-QA behavior may answer an early question correctly using local cues from a single frame, but fail on later questions that require preserving and updating spatial state. 
    This illustrates why isolated question accuracy can overestimate spatial intelligence. 
    (b) By reasoning with the preserved spatial state across frames and question turns, the model can produce answers that remain jointly consistent throughout the chain.
    }
    \label{fig:head}
    \vspace{-0.2cm}
\end{figure}

However, existing data-centric approaches, especially on the training-data side, often rely on collecting and curating large-scale real-world data~\cite{SenseNova, cambrians, robobrain25}, which often incurs substantial collection, annotation, and curation costs. More importantly, current spatial datasets~\cite{SenseNova, cambrians, robobrain25, embspatial, viewspatialbench, vsi-bench}, whether for training or evaluation, typically treat spatial reasoning as a set of independent question-answer (QA) instances. Since models can answer them in isolation without maintaining a coherent understanding of the space, this leaves room for shortcut-based answering (Fig.~\ref{fig:head}(a)) through commonsense priors, local cues, or dataset-specific regularities~\cite{nosense}, thus not only making evaluation less faithful to true spatial intelligence, but also providing only limited supervision during training.

At the core of these limitations is the lack of a structured formulation of persistent spatial state. In real-world embodied settings, spatial reasoning is rarely a collection of isolated questions. Instead, it requires a model to accumulate observations over time, maintain a coherent understanding of the environment, and answer multiple interdependent queries grounded in the same underlying spatial context~\cite{video-msr}, as shown in Fig.~\ref{fig:head}(b). 

Based on this insight, we introduce \texttt{ChainSpace}, a chained-reasoning paradigm that moves beyond independent QA and organizes spatial reasoning as a multi-round process grounded in persistent spatial state. The core principle lies  in organizing multiple questions as a logically constrained process, where later questions depend on earlier spatial constraints and the full answer sequence must remain jointly consistent. Under this paradigm, we instantiate both evaluation and learning. For evaluation, we construct \texttt{ChainSpace-Bench}, a manually annotated real-world multi-round benchmark with a \emph{Chain-Aware Metric} for measuring chain-level consistency beyond isolated question accuracy. 
For learning, we propose \texttt{ChainSpace-Pipeline}, which generates simulator-based chained supervision from embodied trajectories. 
Experiments show that \texttt{ChainSpace-Bench} exposes chain-level failures missed by isolated question accuracy. With only 50K simulator-generated chained samples, models trained by \texttt{ChainSpace-Pipeline} achieve the best performance among open-source models on \texttt{ChainSpace-Bench}, outperform spatial models trained with hundreds of thousands to millions of real-world samples, and transfer competitively to external spatial intelligence benchmarks.
Our contributions are summarized as
\begin{itemize}
    \item We introduce \texttt{ChainSpace}, a chained-reasoning paradigm for spatial intelligence, which organizes spatial questions into logically constrained and jointly consistent chains grounded in persistent spatial state, making shortcut-based answering substantially harder.
    
     \item We instantiate \texttt{ChainSpace} for evaluation through \texttt{ChainSpace-Bench}, a manually annotated real-world multi-round benchmark, together with a \emph{Chain-Aware Metric} for measuring chain-level spatial reasoning beyond isolated per-question accuracy.

    \item We instantiate \texttt{ChainSpace} for learning through \texttt{ChainSpace-Pipeline}, a simulator-based data construction pipeline that generates chained supervision from embodied trajectories, providing more informative supervision for learning spatial intelligence.

    \item Experiments show that \texttt{ChainSpace-Bench} exposes chain-level failures missed by isolated question accuracy. With only 50K simulator-generated chained samples, \texttt{ChainSpace-Pipeline} achieves the best performance among open-source models on \texttt{ChainSpace-Bench}, outperforming spatial models trained with hundreds of thousands to millions of real-world samples, and transfers competitively across multiple external spatial intelligence benchmarks.
\end{itemize}

\section{Related Work}
\label{Related Work}

Existing efforts to evaluate and improve spatial intelligence in foundation models are largely built upon isolated supervision signals. 
For evaluation, benchmarks such as VSI-Bench~\cite{vsi-bench}, ViewSpatial-Bench~\cite{viewspatialbench}, and MMSI-Bench~\cite{MMSI-Bench} evaluate spatial reasoning limitations in multimodal models from different aspects. However, because these benchmarks typically score questions independently, they provide limited evidence of whether a model can preserve, update, and reuse spatial state across related queries.  As a result, isolated accuracy may overestimate spatial intelligence, since models can answer individual questions using local cues, commonsense priors, or dataset-specific regularities~\cite{nosense}. A similar issue appears in training data. 
Recent data-centric efforts, such as SenseNova-SI~\cite{SenseNova}, Cambrian-S~\cite{cambrians}, and RoboBrain~\cite{robobrain25}, show that scaling spatial supervision improves model capability, but their samples are still largely organized as independent QA pairs. 
Such supervision provides limited pressure for models to preserve and update spatial constraints across steps, and may favor local pattern matching over reusable spatial reasoning. In contrast, \texttt{ChainSpace} organizes spatial data as logically dependent and jointly consistent chains. \texttt{ChainSpace-Bench} enables chain-level evaluation beyond isolated accuracy, while \texttt{ChainSpace-Pipeline} generates chained supervision that requires models to reuse accumulated spatial state. 
Together, they instantiate a data-centric paradigm for more faithful evaluation and more informative learning of spatial intelligence.

\section{Chained Spatial Intelligence}
\label{sec:formulation}

Spatial intelligence in embodied settings is better viewed as \emph{chained reasoning under a shared latent spatial state}. 
A spatial context $x$ (\emph{e.g.}, a video trajectory) induces an underlying spatial state, including object locations, viewpoint changes, room topology, and spatial relations accumulated over time~\cite{vsi-bench}. 
Multiple spatial questions are therefore not isolated labels, but different queries over the same latent state~\cite{MindCube, video-msr}. 
Given a multi-round question set $q_{1:T} = (q_1,\dots,q_T)$ and the corresponding answers $a_{1:T} = (a_1,\dots,a_T)$, the conventional independent-QA formulation implicitly treats prediction as
\begin{equation}
p(a_{1:T}\mid x, q_{1:T}) \approx \prod_{t=1}^{T} p(a_t \mid x, q_t),
\label{eq:independent_qa}
\end{equation}
which assumes that each round can be answered in isolation once the question and observation are given. 
This factorization ignores the structural coupling among answers that should be induced by the same spatial state. 
As a result, a model may achieve high per-question accuracy by exploiting local cues, commonsense priors, or dataset-specific regularities, without maintaining a coherent spatial understanding of the full scene~\cite{nosense}. 

\paragraph{Chained Spatial Reasoning.} In contrast, we formulate spatial reasoning as a chained process. 
Let $s_t$ denote the latent spatial state maintained after round $t$, which conceptually summarizes the spatial constraints established so far. 
Here, $s_t$ is not assumed to be an explicit architectural module, but a representation of the persistent spatial information that a model should preserve, update, and reuse across question turns. 
We model chained spatial reasoning as
\begin{equation}
p(a_{1:T}\mid x, q_{1:T}) = \prod_{t=1}^{T} p(a_t \mid x, q_{\leq t}, s_{t-1}),
\label{eq:chained_reasoning}
\end{equation}
with state transition $s_t = \Phi(s_{t-1}, x, q_t, a_t)$, where later predictions are conditioned on the spatial constraints accumulated from earlier rounds. 
From this perspective, Eq.~\ref{eq:independent_qa} can be viewed as a degenerate case in which the intermediate spatial state is ignored or repeatedly reset, while Eq.~\ref{eq:chained_reasoning} treats the answer sequence as a structured prediction problem under a shared spatial context. This formulation induces two key properties for valid chained spatial reasoning. 
The first is \textbf{dependency}, \emph{i.e.}, at least one later-round answer should require constraints established in previous rounds, rather than being reducible to the current question alone. 
The second is \textbf{consistency}, \emph{i.e.}, the full answer tuple $a_{1:T}$ should be jointly valid under the same spatial context. 
Let $\mathcal{C}(x,q_{1:T})$ denote the set of answer chains that satisfy the spatial constraints induced by $x$ and $q_{1:T}$. 
A valid prediction should satisfy
\begin{equation}
a_{1:T} \in \mathcal{C}(x,q_{1:T}),
\end{equation}
rather than merely maximizing isolated correctness for each $a_t$.  This formulation motivates our evaluation and training design.
For evaluation, isolated per-question accuracy only measures marginal correctness and may miss whether answers satisfy pairwise and full-chain spatial constraints.
For training, Eq.~\ref{eq:chained_reasoning} suggests that supervision should preserve cross-round dependencies, requiring later rounds to reuse and update spatial constraints established earlier.
Thus, \texttt{ChainSpace} provides a paradigm for organizing spatial evaluation and supervision around shared-state, logically constrained answer chains.
\texttt{ChainSpace-Bench} and \texttt{ChainSpace-Pipeline} are two concrete instantiations of this paradigm, for evaluation and learning respectively.

\paragraph{Why chained spatial reasoning reduces shortcut answering.}
Since each answer is optimized and scored marginally, a shortcut predictor can produce a plausible local answer from $(x,q_t)$ without ensuring that the full answer tuple is spatially valid.
In other words, independent QA does not explicitly penalize arbitrary answer combinations in $\mathcal{A}^T$, so correct early answers do not necessarily constrain later predictions. Chained reasoning instead turns spatial reasoning into constrained answer-chain prediction.
For a $T$-round chain with answer space $\mathcal{A}$, valid predictions are restricted to $\mathcal{C}(x,q_{1:T}) \subseteq \mathcal{A}^T$, where cross-round constraints eliminate locally plausible but globally inconsistent answer combinations.
Thus, state-agnostic shortcuts become less viable, since later answers must remain compatible with spatial constraints established earlier.


\section{ChainSpace-Bench}
\label{Bench}

\subsection{Overview}
Following the formulation in Sec.~\ref{sec:formulation}, we introduce \texttt{ChainSpace-Bench} to provide a faithful benchmark for evaluating spatial intelligence. The core distinction of \texttt{ChainSpace-Bench} from prior spatial intelligence benchmarks~\cite{vsi-bench, embspatial, viewspatialbench, MMSI-Bench, MindCube, 3dsrbench, SITE} lies in its chained multi-round reasoning. We organize each question set as a logically constrained reasoning process, where later questions depend on the spatial constraints established by earlier ones and the full chain must remain jointly consistent. \texttt{ChainSpace-Bench} contains 326 real-world videos and 1,304 3-rounds QA pairs. To ensure benchmark reliability, all annotations are manually verified under a strict quality-control protocol. We carefully remove erroneous labels, refine ambiguous cases, and only retain samples that reach agreement within the annotation team. More benchmark details can be found in the appendix.

\subsection{Benchmark Construction}

\paragraph{Data Collection.}
We build \texttt{ChainSpace-Bench} from RoomTour3D~\cite{RoomTour3D}, which provides continuous first-person room-tour videos collected from real indoor environments.
These videos contain natural temporal continuity, diverse room layouts, rich object arrangements, and multi-room trajectories, making them suitable for evaluating spatial understanding across time.
Before annotation, we manually filter out videos with overly complex spatial structures, poor camera viewpoints, or insufficient visual coverage that prevents reliable annotation of spatial relations. 

\begin{figure}[t]
    \centering
    \includegraphics[width=1.0\textwidth]{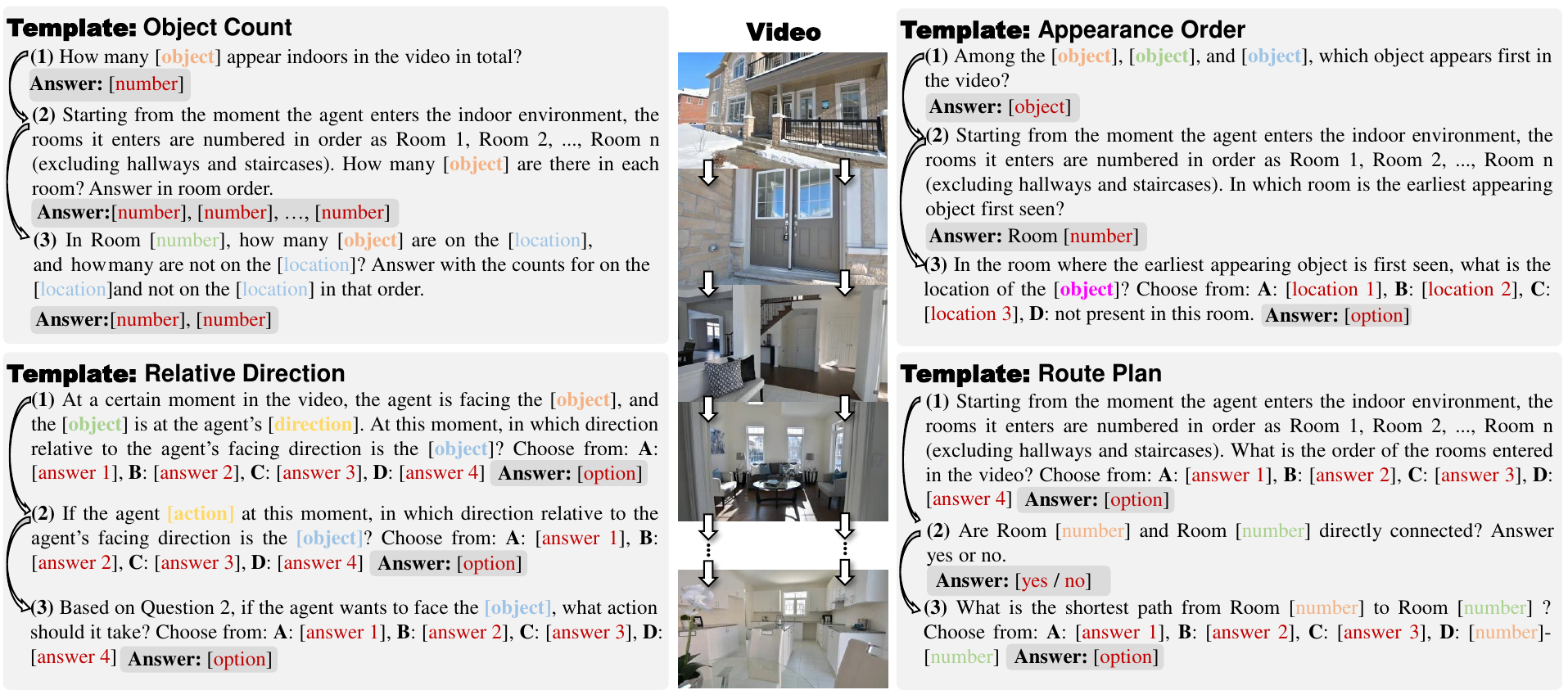}
    \caption{
    Illustration of our task templates.
    Each trajectory is instantiated into four 3-round question templates, \emph{i.e.}, \emph{object count}, \emph{appearance order}, \emph{relative direction}, and \emph{route plan}. All templates follow the same chained design principle, where later questions depend on the spatial constraints established by earlier ones.
    For \emph{object count}, the chain goes from global counting to room-wise decomposition and then to finer-grained counting within a specific room.
    For \emph{appearance order}, it goes from earliest object identification to first-appearance room localization and then to object-location reasoning within that room.
    For \emph{relative direction}, it goes from direction perception to hypothetical viewpoint change and then to action prediction.
    For \emph{route plan}, it goes from room visitation sequence to room adjacency and then to shortest-path reasoning.
    }
    \label{fig:template}
    \vspace{-0.2cm}
\end{figure}

\paragraph{Question-Answer Annotation.} Spatial intelligence involves a broad range of abilities, including counting, spatial relations, navigation, and size or distance estimation~\cite{SenseNova,vsi-bench,viewspatialbench}. 
For each selected video, we annotate four spatial reasoning tasks, \emph{i.e.}, object count, appearance order, relative direction, and route plan. 
We choose these 4 tasks because they are reliably annotatable on real-world videos and naturally support chained logical dependencies across questions. 
In contrast, size- and distance-centric tasks are more sensitive to viewpoint changes, making precise human annotation harder on real-world trajectories~\cite{RSA,viewspatialbench}. To ensure consistency and reduce annotation ambiguity, we design a dedicated question template for each task type. Each template forms a 3-round QA chain, where the questions are progressively organized to reflect the intended logical dependency within the task. We adopt a 3-round design because it provides the minimal structure needed to move beyond isolated QA and simple pairwise dependency, while still remaining practical for reliable human annotation and review. The question templates for the four task types are illustrated in Fig.~\ref{fig:template}.

\paragraph{Annotation Quality.} We adopt a two-stage quality-control process that combines automatic rule-based filtering with human-in-the-loop review. We first apply rule-based checks to remove QA chains with invalid answer formats, inconsistent object or room references, ambiguous options, or violations of the intended cross-round dependencies. The reviewer team then continuously monitors the annotation progress, examines the correctness of labeled QA chains, and provides timely feedback to the annotators whenever ambiguities or inconsistencies arise. Annotations are only retained when agreement is reached within the team. 

\subsection{Details of \texttt{ChainSpace-Bench}}

\begin{figure}[!htb]
    \centering
    \includegraphics[width=0.9\textwidth]{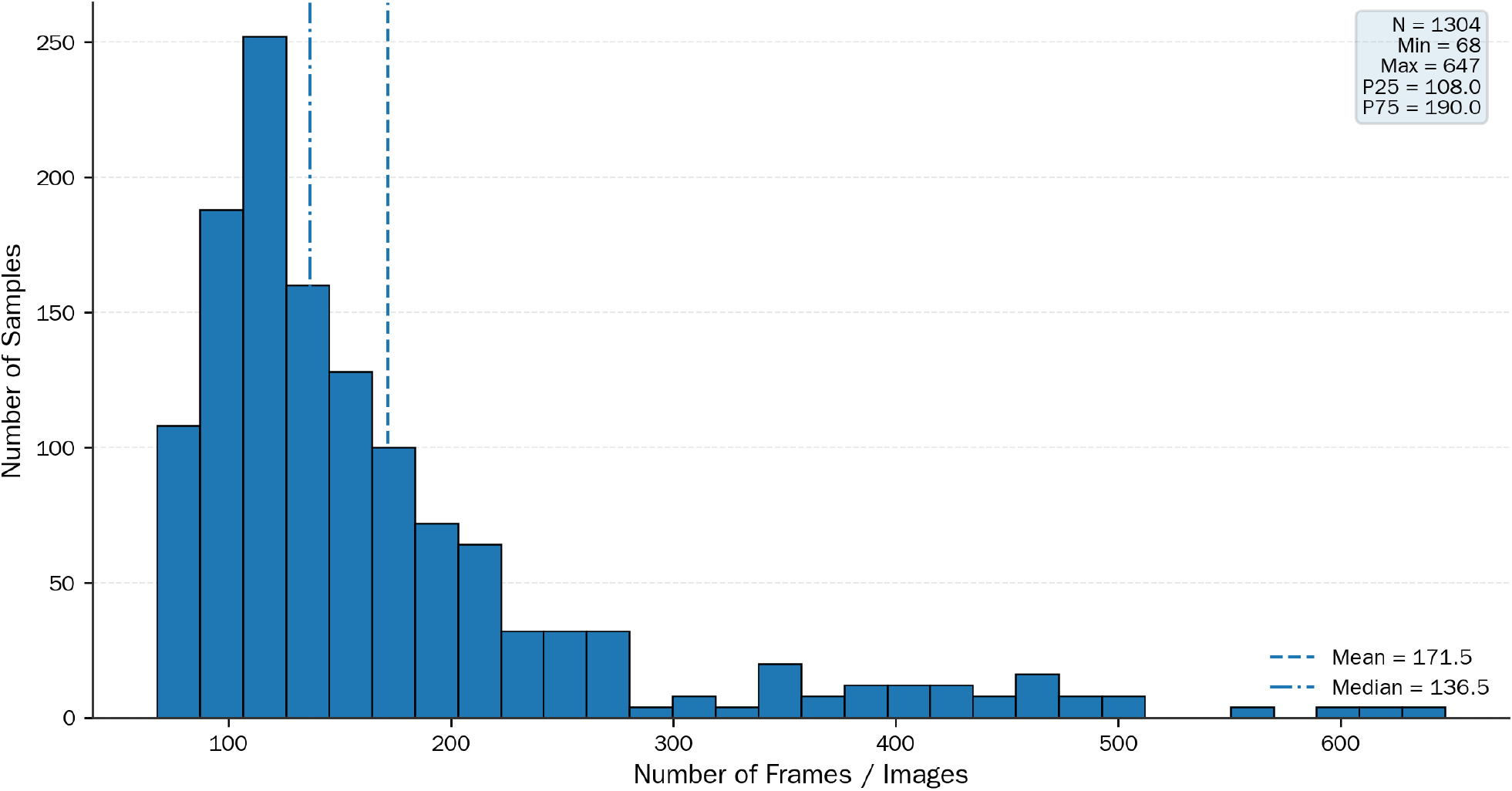}
    \caption{Frame count distribution of \texttt{ChainSpace-Bench}. The benchmark covers trajectories with diverse lengths, with most samples concentrated in the moderate range and a long tail of substantially longer videos.}
    \label{fig:profile2}
\end{figure}

\begin{figure}[!htb]
    \centering
    \includegraphics[width=0.9\textwidth]{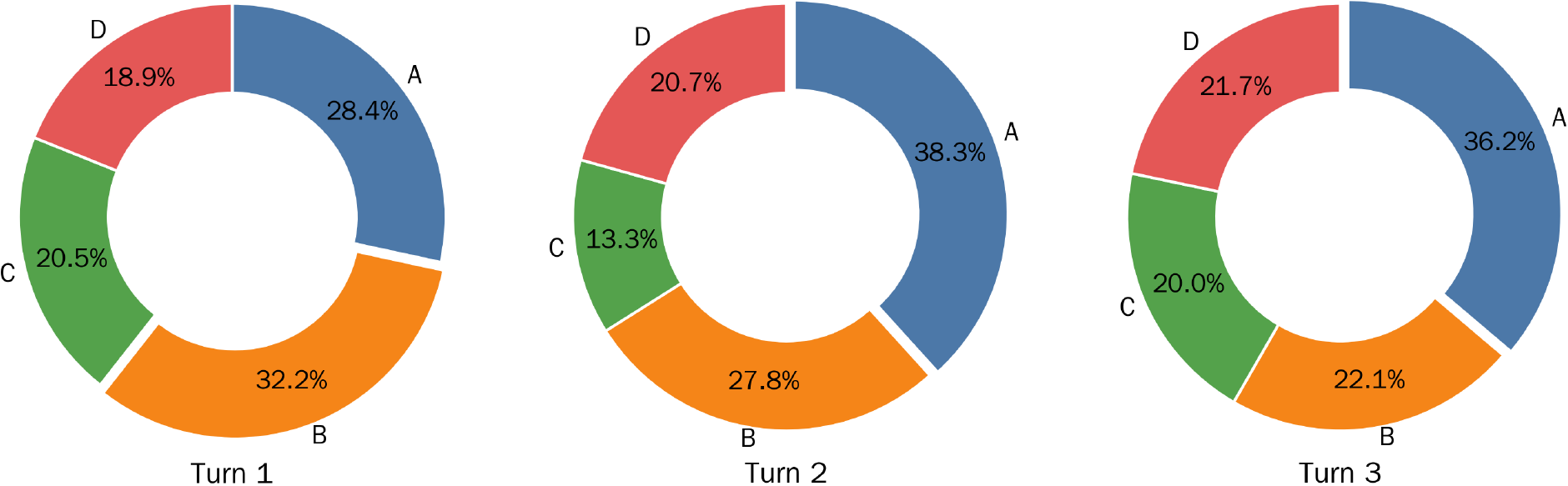}
    \caption{The A/B/C/D answer distribution in \texttt{ChainSpace-Bench} by turn. The option labels are reasonably balanced across rounds, reducing the possibility that models can perform well by exploiting simple answer-frequency bias.}
    \label{fig:profile1}
\end{figure}

\paragraph{Dataset profile of \texttt{ChainSpace-Bench}.}
\texttt{ChainSpace-Bench} exhibits substantial diversity in both temporal complexity and answer space. As shown in Fig.~\ref{fig:profile2}, the benchmark covers trajectories with varied lengths, with most samples concentrated in a moderate frame range and a long tail of substantially longer videos, indicating that evaluation is not restricted to short or temporally simple cases. In addition, Fig.~\ref{fig:profile1} shows that the A/B/C/D answer distribution remains reasonably balanced across all three rounds, reducing the possibility that models can achieve strong performance by exploiting simple option-frequency bias. Moreover, the benchmark is also well balanced in task-specific aspects. For action prediction, different actions are distributed relatively evenly, with turns toward different directions each accounting for roughly 10\% of the answers. For room-identification questions, the answer distribution is led by \texttt{Room 1} and \texttt{Room 2}, which account for 33.2\% and 27.5\% respectively, while the remaining rooms are more evenly distributed. For object-location questions, different location categories are all distributed at roughly 2\%, suggesting that no single location prior dominates the benchmark. Finally, object categories are also diverse. Although \emph{TV}, \emph{lamp}, and \emph{refrigerator} are the most frequent categories, each accounts for only about 3\% of all object mentions. Overall, these statistics suggest that \texttt{ChainSpace-Bench} is not dominated by a few trivial priors, but instead provides a relatively diverse and balanced testbed for chained spatial reasoning.

\subsection{Chain-Aware Metric}

To align evaluation with the principle of \texttt{ChainSpace}, we propose a \emph{Chain-Aware Metric}. Instead of scoring questions independently~\cite{vsi-bench,MMSI-Bench,viewspatialbench}, it rewards both local correctness and pairwise/full-chain consistency. Because \texttt{ChainSpace-Bench} contains manually verified cross-round dependencies, correctness patterns indicate whether predictions follow the intended spatial chain. The metric is defined as
\begin{equation}
R_{\mathrm{chain}}(\mathbf{s})=
\sum_{t=1}^{3}\alpha_t s_t
+
\sum_{i < j}\beta_{ij} s_i s_j
+
\gamma s_1 s_2 s_3,
\label{chain-aware-metric}
\end{equation}
where $s_t \in \{0,1\}$ denotes whether the answer at round $t$ is correct and $\mathbf{s}=(s_1,s_2,s_3)$.
The $\alpha$ terms measure local correctness, the $\beta$ terms reward pairwise cross-round consistency, and $\gamma$ captures full-chain consistency.
The score is normalized by the maximum attainable value.
In our experiments, we set $\alpha_1=\alpha_2=\alpha_3=0.5$, $\beta_{12}=2.0$, $\beta_{23}=1.5$, $\beta_{13}=1.25$, and $\gamma=3.0$.
We assign smaller weights to isolated correctness and larger weights to chained terms, so that the metric emphasizes whether the full spatial reasoning chain is established rather than being dominated by per-question accuracy.
Among pairwise terms, $\beta_{12}$ is largest because $(s_1,s_2)$ captures the most direct local dependency, while $\beta_{23}$ and $\beta_{13}$ are lower to reflect incomplete or less reliable chain evidence without the full intermediate reasoning path.
Finally, $\gamma$ receives the largest weight to reward fully correct and jointly consistent chains.
We further ablate these hyper-parameters in Table~\ref{tab:weight_ablation_full} to show the robustness of weight choices.

\section{ChainSpace-Pipeline}
\label{Pipeline}


\subsection{Overview}

Following the principle in Sec.~\ref{sec:formulation}, we propose \texttt{ChainSpace-Pipeline}, a simulator-based chain-structured supervision generation framework for spatial intelligence training. 
We use ProcTHOR~\cite{procthor} and AI2-THOR~\cite{ai2thor} to generate diverse indoor scenes and collect continuous exploration trajectories through automatic embodied navigation. 
For each trajectory, we construct 4 task instances, corresponding to object count, appearance order, relative direction, and route plan. 
Each instance is instantiated as a 3-round chained template from trajectory-specific spatial states and metadata, so that later rounds depend on spatial constraints established earlier. Unlike prior data-construction pipelines that largely organize supervision as independent QA~\cite{SenseNova, cambrians, SIMS-VSI}, \texttt{ChainSpace-Pipeline} preserves the dependency structure of spatial reasoning. This turns simulator metadata into chained supervision that requires models to reuse and update accumulated spatial state across rounds, providing a more structured training signal for learning spatial intelligence.

\subsection{Pipeline Construction}

\paragraph{Trajectory and Scene Generation}
To construct training data at scale, we adopt a simulator-based pipeline. This enables low-cost generation of a large amount of data, which is difficult to obtain through manual annotation on real-world videos. Moreover, since the model will be evaluated on real-world benchmarks, this setup helps verify that performance gains come from the proposed chained design principle itself, rather than from additional training on benchmark-specific data. To support such a pipeline, the simulator needs to generate diverse multi-room indoor scenes with rich object layouts and provide precise spatial states together with object-level metadata for automatic question construction. We therefore use ProcTHOR~\cite{procthor} to generate indoor houses, and AI2-THOR~\cite{ai2thor} to collect continuous exploration trajectories through automatic embodied navigation. During exploration, the agent traverses reachable positions while we record RGB observations, agent poses, room transitions, visible objects, and house-level structural metadata. As a result, the generated trajectories exhibit diverse spatial layouts, realistic navigation patterns, and continuous spatial context over time, making them well suited for constructing logically constrained multi-round supervision.

\paragraph{Chained Question-Answer Construction.}
For each scene and exploration trajectory, we construct four task instances, one for each task type (\emph{e.g.}, object count, appearance order, relative direction, and route plan). Each instance is formulated as a 3-round chained question template (Fig.~\ref{fig:template}) instantiated from trajectory-specific spatial states and metadata, where later questions depend on earlier spatial constraints. Fig.~\ref{fig:template} shows the canonical template form. In the actual training data, we vary the phrasing so that supervision is tied to the chained reasoning structure rather than to fixed linguistic patterns.

\emph{Why choose these 4 tasks?} We focus on these 4 tasks because they cover representative spatial abilities, while naturally supporting explicit cross-round dependencies. Other spatial abilities, such as size- and distance-centric reasoning, are also important but less suitable for our current pipeline, as they require precise numerical estimation and are more sensitive to simulator-to-real discrepancies~\cite{SenseNova}. Heavy training on such synthetic numerical signals may further bias the model's depth-related priors and hurt generalization~\cite{atapour2018realtime,maximov2020focus}.

\emph{Why choose 3-round QA chain?} We adopt a 3-round design because it is sufficient to create meaningful chained supervision. The first round establishes a spatial constraint, the second refines it, and the third requires further reasoning under the accumulated context~\cite{turnbenchms}. This yields non-trivial cross-round dependencies while remaining stable for large-scale automatic construction, whereas longer chains are more likely to introduce ambiguity, noise, and error propagation~\cite{turnbenchms,errorpropcot}. 

\paragraph{Quality Control and Data Filtering.}
We ensure training data quality through rule-based filtering followed by human inspection. For each generated QA chain, we first apply automatic checks to remove invalid answer formats, inconsistent object or room references, ambiguous options, violated cross-round dependencies, and trajectories with excessively long videos (more than 100 frames). We then manually review a validation subset of 100 trajectories (400 chained QA samples) using the same reviewer team as in Sec.~\ref{Bench}. Full-scale data generation is finalized only after this subset passes human inspection, and the remaining filtered samples constitute the final training set.

\section{Experiment}
\label{Experiment}

\subsection{Experimental Setup}
\label{sec:setup}
\paragraph{Datasets.}
For training, we use the simulation data generated by \texttt{ChainSpace-Pipeline}. For evaluation, we use two groups of benchmarks. Our annotated \texttt{ChainSpace-Bench} is used to directly test the proposed chained design principle, while eight existing spatial intelligence benchmarks, including VSI-Bench~\cite{vsi-bench}, MMSI-Bench~\cite{MMSI-Bench}, MindCube~\cite{MindCube}, ViewSpatial~\cite{viewspatialbench}, EmbSpatial~\cite{embspatial}, SITE-Bench~\cite{SITE}, 3DSRBench~\cite{3dsrbench}, and NavBench~\cite{navbench} used to assess the generalization of the learned spatial reasoning ability.

\paragraph{Evaluation Metrics}
On \texttt{ChainSpace-Bench}, we report both per-question accuracy and our proposed \emph{Chain-Aware Metric}. Per-question accuracy measures local correctness at each round, while the Chain-Aware Metric further evaluates logical consistency across the full reasoning chain. Besides, we report Full-chain Exact Match as a strict unweighted chain-level metric, ensuring that the observed gain does not depend on a particular weighting scheme. For existing spatial intelligence benchmarks, we follow their original evaluation protocols. Specifically, MMSI-Bench~\cite{MMSI-Bench}, MindCube~\cite{MindCube}, ViewSpatial~\cite{viewspatialbench}, and EmbSpatial~\cite{embspatial} are evaluated using accuracy. VSI-Bench~\cite{vsi-bench} reports exact-match accuracy for multiple-choice questions and Mean Relative Accuracy (MRA) for numerical questions. For the end-to-end embodied navigation evaluation on NavBench~\cite{navbench}, we report \emph{Success Rate} (SR) and \emph{Success weighted by Path Length} (SPL) under the easy, medium, and hard difficulty levels. We additionally report \emph{Exec. Avg.}, defined as the average of SR and SPL across the three difficulty levels, to summarize overall embodied-navigation performance.

\paragraph{Implementation Details}
We use Qwen3-VL-8B-instruct~\cite{qwen3} as the main base model. InternVL3-8B~\cite{internvl} is also used to evaluate our generalization ability. The models are trained on 8 NVIDIA H20 GPUs with full fine-tuning, where the ViT encoder is frozen. We adopt bfloat16 training with FlashAttention, gradient checkpointing, and DeepSpeed ZeRO-2 for memory-efficient optimization. For training on \texttt{ChainSpace-Pipeline}, we use 1 epoch, a per-device batch size of 1, gradient accumulation steps of 4, and a learning rate of $1\times10^{-7}$ with a warmup ratio of 0.05. The maximum sequence length is set to 32768. We also enable DFT loss during training.

\subsection{Main Results}

\paragraph{Faithful reflection of spatial intelligence.}
As shown in Table~\ref{tab:chainspace_main_result}, 
\texttt{ChainSpace-Bench} \emph{more faithfully reflects spatial intelligence.} RoboBrain2.5~\cite{robobrain25} improves the Avg. Question Accuracy over the Qwen3-VL-8B~\cite{qwen3} baseline after large-scale spatial training, yet its Chain-Aware Score decreases. This mismatch indicates that isolated question accuracy alone may overestimate a model's spatial ability~\cite{nosense}. The proprietary-model results also suggest that \texttt{ChainSpace-Bench} is non-trivial even for strong frontier systems, while the larger differences under the Chain-Aware score indicate that our metric provides a discriminative power beyond isolated question accuracy.

\begin{table}[!htb]
\centering

\footnotesize

\caption{
Results on \texttt{ChainSpace-Bench}. Open-source models and our method are all built on Qwen-family (Qwen2.5-VL-7B-Instruct~\cite{qwen25} and Qwen3-VL-8B-Instruct~\cite{qwen3}) for fair comparison. Existing baselines are trained with different data scales, including CamBrain-S ($\sim$590K), RoboBrain2.0 ($\sim$5.3M), RoboBrain2.5 ($\sim$12.4M), and SenseNova-SI ($\sim$8M), while our \texttt{ChainSpace-Pipeline} uses only $\sim$50K simulator-generated samples. The text in \textcolor{red}{red} and \textcolor{blue}{blue} denotes the best and second-best results among open-source models.
}

\begin{tabular}{lcc}
\toprule
\textbf{Model} & \textbf{Avg. Question Accuracy} & \textbf{Avg. Chain-Aware Score} \\
\rowcolor{blue!5}\multicolumn{3}{l}{\textbf{Proprietary Models}} \\
Seed-2.0-Pro~\cite{seed2.0}     & 0.4289 & 0.2722 \\
GPT-5~\cite{gpt5}     & 0.4825 & 0.3262  \\
Gemini-3.1-Pro~\cite{gemini3}     & 0.3453 & 0.1882 \\
\midrule
\rowcolor{blue!5}\multicolumn{3}{l}{\textbf{Open-source General Models}} \\
Qwen2.5-VL-7B-Instruct~\cite{qwen25}     & 0.2344 & 0.0819 \\
Qwen3-VL-8B-Instruct~\cite{qwen3}     & 0.3344 & 0.1522 \\ \midrule
\rowcolor{blue!5}\multicolumn{3}{l}{\textbf{Open-source Spatial Intelligence Models}} \\
CamBrain-S-7B~\cite{cambrians}          & 0.1824 & 0.0674 \\
RoboBrain2.0-7B~\cite{robobrain2.0}          & 0.2873 & 0.1098  \\
RoboBrain2.5-8B~\cite{robobrain25}          & 0.3384 & 0.1339 \\ 
SenseNova-SI-8B~\cite{SenseNova}          & \textcolor{blue}{0.3418} & \textcolor{blue}{0.1555} \\ \midrule
\rowcolor{pink!15}(\textbf{Ours}) \texttt{ChainSpace-Pipeline}-8B               & \textbf{\textcolor{red}{0.3510}} & \textbf{\textcolor{red}{0.1687}} \\
\bottomrule
\end{tabular}
\label{tab:chainspace_main_result}
\vspace{-0.2cm}
\end{table}

\begin{table}[!htb]
\centering
\footnotesize
\setlength{\tabcolsep}{4.8pt}
\renewcommand{\arraystretch}{1.12}
\caption{
Paired comparison against the same baseline (Qwen3-VL-8B-Instruct~\cite{qwen3}) on \texttt{ChainSpace-Bench}.
Each cell reports the mean paired difference $\Delta$, the 95\% paired-bootstrap confidence interval, and the paired-bootstrap $p$-value in the format
$\Delta$ [CI] ($p$).
\textbf{Bold} indicates statistically significant change ($p<0.05$).
}
\label{tab:paired_vs_baseline_main}
\begin{tabular}{lcc}
\toprule
\textbf{Metric} & (\textbf{Ours}) \texttt{ChainSpace-Pipeline} & \textbf{SenseNova-SI} \\
\midrule
(\textit{Chain-level metrics}) Full-chain Exact Match &
\cellcolor{pink!15}\textbf{+0.0153} {\scriptsize [0.0054, 0.0261] (0.0044)} &
{\scriptsize +0.0038 [-0.0100, 0.0176] (0.6380)} \\

(\textit{Chain-level metrics}) Avg. Chain-Aware Score &
\cellcolor{pink!15}\textbf{+0.0165} {\scriptsize [0.0079, 0.0256] (<0.0001)} &
{\scriptsize +0.0033 [-0.0086, 0.0154] (0.5980)} \\
\midrule
(\textit{Turn-level metrics}) Turn-1 Accuracy &
\cellcolor{pink!15}\textbf{+0.0261} {\scriptsize [0.0146, 0.0383] (<0.0001)} &
{\scriptsize +0.0092 [-0.0146, 0.0330] (0.4720)} \\

(\textit{Turn-level metrics}) Turn-2 Accuracy &
{\scriptsize +0.0107 [-0.0038, 0.0253] (0.1692)} &
\cellcolor{blue!5}\textbf{-0.0215} {\scriptsize [-0.0376, -0.0054] (0.0116)} \\

(\textit{Turn-level metrics}) Turn-3 Accuracy &
{\scriptsize +0.0130 [-0.0077, 0.0345] (0.2416)} &
\cellcolor{blue!5}\textbf{+0.0345} {\scriptsize [0.0084, 0.0606] (0.0108)} \\
\bottomrule
\end{tabular}
\label{tab:paired_vs_baseline_fullcompare}
\vspace{-0.2cm}
\end{table}

\begin{table}[!htb]
\centering
\footnotesize
\caption{
Oracle-history intervention on ChainSpace-Bench.
``Pred'' denotes normal inference using model-generated previous answers, while
``Gold'' replaces previous answers with ground-truth answers during inference.
Relative gains are computed as $(\mathrm{Gold}-\mathrm{Pred})/\mathrm{Pred}$. The text in \textcolor{red}{red} and \textcolor{blue}{blue} denotes the best and second-best results of each setting.
}
\label{tab:oracle_history}
\setlength{\tabcolsep}{5.5pt}
\renewcommand{\arraystretch}{1.12}
\begin{tabular}{llcccc}
\toprule
\textbf{Model} & \textbf{Setting} & \textbf{Avg. Acc.} & \textbf{Q2 Acc.} & \textbf{Q3 Acc.} & \textbf{Chain-Aware} \\
\midrule
\multirow{3}{*}{Qwen3-VL-8B~\cite{qwen3}}
& Pred & 0.3344 & \textcolor{blue}{0.3574} & 0.3497 & 0.1522 \\
& Gold & 0.3563 & \textcolor{blue}{0.3635} & 0.4095 & 0.1589 \\
& Gain & \textcolor{blue}{+6.55\%} & +1.71\% & \textcolor{blue}{+17.10\%} & +4.40\% \\
\midrule
\multirow{3}{*}{SenseNova-SI-8B~\cite{SenseNova}}
& Pred & \textcolor{blue}{0.3418} & 0.3359 & \textcolor{red}{\textbf{0.3842}} & \textcolor{blue}{0.1555} \\
& Gold & \textcolor{blue}{0.3635} & 0.3528 & \textcolor{blue}{0.4325} & \textcolor{blue}{0.1628} \\
& Gain & +6.35\% & \textcolor{red}{\textbf{+5.03\%}} & +12.57\% & \textcolor{blue}{+4.69\%} \\
\midrule
\multirow{3}{*}{(\textbf{Ours}) \texttt{ChainSpace-Pipeline}-50k}
& Pred & \textcolor{red}{\textbf{0.3510}} & \textcolor{red}{\textbf{0.3681}} & \textcolor{blue}{0.3627} & \textcolor{red}{\textbf{0.1687}} \\
& Gold & \textcolor{red}{\textbf{0.3873}} & \textcolor{red}{\textbf{0.3781}} & \textcolor{red}{\textbf{0.4617}} & \textcolor{red}{\textbf{0.1806}} \\
& Gain & \textcolor{red}{\textbf{+10.34\%}} & \textcolor{blue}{+2.72\%} & \textcolor{red}{\textbf{+27.30\%}} & \textcolor{red}{\textbf{+7.05\%}} \\
\bottomrule
\end{tabular}
\vspace{-0.2cm}
\end{table}

\paragraph{Reliable chained spatial reasoning.}
Tables~\ref{tab:chainspace_main_result}, \ref{tab:paired_vs_baseline_fullcompare}, and~\ref{tab:oracle_history} provide a progressive evidence chain for the effectiveness of \texttt{ChainSpace}
\textbf{(1)} \texttt{ChainSpace-Pipeline} \emph{improves chain-level spatial reasoning with much less data.}
As shown in Table~\ref{tab:chainspace_main_result}, our model outperforms open-source spatial models trained on substantially larger real-world datasets, despite using only $\sim$50K simulator-generated samples.
This suggests that organizing supervision under \texttt{ChainSpace} can be more effective than simply scaling independent spatial QA data.
\textbf{(2)} \emph{The improvement is reliable and concentrated on chain-level metrics.}
Table~\ref{tab:paired_vs_baseline_fullcompare} compares different tuned models against the same untuned baseline.
Our method significantly improves both Full-chain Exact Match and Chain-Aware Score, while SenseNova-SI~\cite{SenseNova} does not show significant gains on either chain-level metric.
This indicates that the benefit of our pipeline is not merely better isolated answering, but more complete chained reasoning.
\textbf{(3)} \emph{The gain comes from better use of accumulated spatial state.}
Table~\ref{tab:oracle_history} performs an oracle-history intervention.
Notably, our model obtains the largest gains in Q3 Acc. under correct history, revealing that correct spatial history unlocks substantially stronger downstream reasoning.
This suggests that the main bottleneck of previous trained spatial models is the ability to preserve and propagate correct spatial state across rounds. \texttt{ChainSpace-Pipeline} encourages models to update and reuse accumulated spatial constraints, enabling them to benefit more effectively from correct reasoning history.

\begin{table}[t]
\centering

\caption{
Performance on the enlarged ChainSpace-Bench containing 800 videos and 3,200 chains.
The \textcolor{red}{red} and \textcolor{blue}{blue} denote the best and second-best results, respectively.
}
\label{tab:expanded_benchmark}

\resizebox{0.75\linewidth}{!}{
\begin{tabular}{lcc}
\toprule
\textbf{Model} &
\textbf{Avg. Question Accuracy} &
\textbf{Chain-Aware Score} \\
\midrule

Qwen3-VL-8B-Instruct~\cite{qwen3}
& 0.3348
& 0.1531 \\

SenseNova-SI-8B~\cite{SenseNova}
& \textcolor{blue}{0.3427}
& \textcolor{blue}{0.1562} \\

\textbf{ChainSpace-Pipeline-8B (Ours)}
& \textcolor{red}{\textbf{0.3546}}
& \textcolor{red}{\textbf{0.1690}} \\

\bottomrule
\end{tabular}
}
\end{table}

\begin{table*}[t]
\centering
\caption{
Controlled comparison of different training paradigms on ChainSpace-Bench and external
spatial reasoning benchmarks. All settings use the same backbone. Independent-QA uses the same
videos, questions, answers, supervision content, and training configuration as ChainSpace-Pipeline,
while removing cross-round dependencies. The \textcolor{red}{red} and \textcolor{blue}{blue} denote
the best and second-best results, respectively.
}
\label{tab:independent_qa}
\small

\resizebox{\textwidth}{!}{
\begin{tabular}{lccccc}
\toprule
\textbf{Models}
& \multicolumn{2}{c}{\textbf{ChainSpace-Bench}}
& \textbf{VSI-Bench~\cite{vsi-bench}}
& \textbf{MMSI-Bench~\cite{MMSI-Bench}}
& \textbf{MindCube~\cite{MindCube}} \\
\cmidrule(lr){2-3}

\textbf{Metric}
& \textbf{Avg. Q. Acc}
& \textbf{Avg. Chain-Aware Score}
& \textbf{MRA, Acc}
& \textbf{Acc}
& \textbf{Acc} \\
\midrule

Qwen3-VL-8B-Instruct~~\cite{qwen3}
& \textcolor{blue}{0.3344}
& \textcolor{blue}{0.1522}
& 57.9
& \textcolor{blue}{31.1}
& 29.4 \\

Independent-QA Control (8B)
& 0.3236
& 0.1482
& \textcolor{blue}{58.1}
& 30.4
& \textcolor{blue}{30.2} \\

\textbf{ChainSpace-Pipeline-8B (Ours)}
& \textcolor{red}{\textbf{0.3510}}
& \textcolor{red}{\textbf{0.1687}}
& \textcolor{red}{\textbf{58.8}}
& \textcolor{red}{\textbf{31.8}}
& \textcolor{red}{\textbf{31.1}} \\

\bottomrule
\end{tabular}
}
\end{table*}

\textbf{(4)} \emph{The advantage remains stable.} We further expand the benchmark from 326 to \textbf{800 videos} and from 1,304 to \textbf{3,200 chains}, corresponding to a \textbf{145.4\% increase} in evaluation scale. As shown in Table~\ref{tab:expanded_benchmark}, the enlarged benchmark preserves the main performance trend. ChainSpace-Pipeline-8B consistently outperforms both Qwen3-VL-8B-Instruct~\cite{qwen3} and SenseNova-SI-8B~\cite{SenseNova} in terms of average question accuracy and Chain-Aware Score. In particular, the Chain-Aware Score of ChainSpace-Pipeline-8B changes only marginally from 0.1687 on the original benchmark to 0.1690 on the enlarged benchmark. These results indicate that the observed advantage remains stable under a substantially larger evaluation set and is unlikely to be an artifact of the original benchmark scale.

\paragraph{Effect of Chained Supervision.}
To isolate the effect of chained supervision, we construct an \textit{Independent-QA} control by converting each ChainSpace chain into independent samples, where the preceding ground-truth answer is explicitly inserted into the subsequent question. The control uses the same backbone, videos, underlying questions and answers, number of QA turns, supervision content, and training configuration as ChainSpace-Pipeline, with the cross-round dependency structure being the only difference. As shown in Table~\ref{tab:independent_qa}, Independent-QA achieves 0.3236 average question accuracy and 0.1482 Chain-Aware Score on ChainSpace-Bench, both below the untuned backbone, whereas ChainSpace-Pipeline improves them to 0.3510 and 0.1687. This controlled comparison shows that simply training on the same videos and QA supervision does not yield the improvement. Preserving cross-round dependencies is essential. Moreover, ChainSpace-Pipeline consistently outperforms the matched Independent-QA control on all three external spatial reasoning benchmarks. These results further demonstrate that the benefit of chained supervision extends beyond ChainSpace-Bench and transfers to external spatial reasoning tasks.

\paragraph{Efficient and generalizable spatial intelligence learning.}
Table~\ref{tab:main_results} evaluates the generalization ability of models trained with \texttt{ChainSpace-Pipeline}. \textbf{(1)} \emph{The learned capability transfers beyond} \texttt{ChainSpace-Bench}. Our models remain competitive and often improve over their corresponding untuned backbones on multiple external spatial intelligence benchmarks.
\textbf{(2)} \emph{Our supervision is data-efficient.} With only 50K simulator-generated chained samples, our Qwen3-VL-8B-Instruct model achieves performance comparable to SenseNova-SI-200K~\cite{SenseNova}, despite using substantially less training data. Since \texttt{ChainSpace-Pipeline} is built from purely simulator-generated data and remains distributionally distinct from all evaluation benchmarks, these results provide additional evidence that our design principle contributes to spatial intelligence learning beyond the training distribution. Moreover, compared with prior approaches, our method remains comparable with much smaller training data, showing that better-organized supervision can be more effective than simply increasing independent QA samples. 
\textbf{(3)} \emph{The gain is not limited to a single backbone.} 
Consistent improvements on both Qwen3-VL-8B-Instruct~\cite{qwen3} and InternVL3-8B~\cite{internvl} suggest that the proposed chained supervision principle is broadly useful for training spatial intelligence.

\begin{table}[!htb]
\centering
\caption{
Comparison between proprietary models, open-source general models, and open-source spatial intelligence models across five spatial intelligence benchmarks. We report our 50K and 150K settings to highlight data efficiency. MindCube* denotes MindCube-Tiny. 
}
\setlength{\tabcolsep}{3pt}
\renewcommand{\arraystretch}{1.15}
\resizebox{\textwidth}{!}{
\begin{tabular}{lccccc}
\toprule
\textbf{Models} & \textbf{VSI-Bench~\cite{vsi-bench}} & \textbf{MMSI-Bench~\cite{MMSI-Bench}} & \textbf{MindCube*~\cite{MindCube}} & \textbf{ViewSpatial~\cite{viewspatialbench}} & \textbf{EmbSpatial~\cite{embspatial}} \\
\midrule
\textbf{Metric} & \textbf{MRA, Acc} & \textbf{Acc} & \textbf{Acc} & \textbf{Acc} & \textbf{Acc} \\
\midrule

\textbf{Random Choice} & 34.0 & 25.0 & 33.0 & 26.3 & - \\
\midrule

\rowcolor{blue!5} \multicolumn{6}{l}{\textbf{Proprietary Models}} \\
Seed-1.6-2025-06-15~\cite{seed1.6} & 49.9 & 38.3 & 48.7 & 43.8 & - \\
Gemini-2.5-Pro-2025-06~\cite{gemini} & 53.5 & 38.0 & 57.6 & 46.0 & 78.9 \\
Grok-4-2025-07-09~\cite{grok4} & 47.9 & 37.8 & 63.5 & 43.2 & 75.7 \\
GPT-5-2025-08-07~\cite{gpt5} & 55.0 & 41.8 & 56.3 & 45.5 & 81.6 \\
Gemini-3-Pro-Preview~\cite{gemini3} & 52.5 & 45.2 & 70.8 & 50.3 & - \\
\midrule

\rowcolor{blue!5}\multicolumn{6}{l}{\textbf{Open-source General Models}} \\
Bagel-7B-MoT~\cite{bagel} & 31.4 & 31.0 & 34.7 & 41.3 & 73.1 \\
Qwen2.5-VL-7B-Instruct~\cite{qwen25} & 32.3 & 26.8 & 36.0 & 36.8 & - \\
Qwen3-VL-8B-Instruct~\cite{qwen3} & 57.9 & 31.1 & 29.4 & 42.2 & 77.7 \\
InternVL3-8B~\cite{internvl} & 42.1 & 28.0 & 41.5 & 38.6 & 76.4 \\
\midrule

\rowcolor{blue!5}\multicolumn{6}{l}{\textbf{Open-source Spatial Intelligence Models}} \\
SpatialLadder-3B~\cite{spatialladder} & 44.8 & 27.4 & 43.4 & 39.8 & - \\
Spatial-MLLM-4B~\cite{wu2025spatialmllm} & 46.3 & 26.1 & 33.4 & 34.6 & - \\
SpaceR-7B~\cite{spacer} & 41.5 & 27.4 & 37.9 & 35.8 & 66.9 \\
ViLaSR-7B~\cite{vilasr} & 44.6 & 30.2 & 35.1 & 35.7 & 67.3 \\
SenseNova-SI-100K$_{\text{Qwen3-VL-8B}}$~\cite{SenseNova} & 56.4  & 31.4 & 30.9 & 42.8 & 79.1 \\
SenseNova-SI-200K$_{\text{Qwen3-VL-8B}}$~\cite{SenseNova} & 58.7 & 30.9  & 29.4 & 43.5 & 79.1 \\

\midrule

\rowcolor{blue!5}\multicolumn{6}{l}{\textbf{Ours}} \\
\rowcolor{pink!15}Qwen3-VL-8B$_{\text{Ours-50K}}$
& 58.8(\textcolor{cyan!60!black}{+1.5\%})
& 31.8(\textcolor{cyan!60!black}{+2.3\%})
& 31.1(\textcolor{cyan!60!black}{+7.1\%})
& 43.0(\textcolor{cyan!60!black}{+2.1\%})
& 78.1(\textcolor{cyan!60!black}{+0.5\%})\\
\rowcolor{pink!15}InternVL3-8B$_{\text{Ours-150K}}$
& 43.6(\textcolor{cyan!60!black}{+3.6\%})
& 28.4(\textcolor{cyan!60!black}{+1.4\%})
& 44.5(\textcolor{cyan!60!black}{+7.2\%})
& 46.8(\textcolor{cyan!60!black}{+21.2\%})
& 76.2(\textcolor{cyan!60!black}{-0.2\%})\\
\bottomrule
\end{tabular}
}
\label{tab:main_results}

\end{table}

We further evaluate our method on two additional spatial intelligence benchmarks that are distributionally distinct from our simulator-generated training data. SITE-Bench~\cite{SITE} is a broad spatial intelligence benchmark covering multiple sub-skills, including counting and existence, object localization, 3D information understanding, multi-view reasoning, spatial relations, and movement/navigation. 3DSRBench~\cite{3dsrbench} focuses on 3D spatial reasoning under several transformed evaluation settings that test robustness to viewpoint and orientation changes.

\begin{table}[!htb]
\centering
\caption{Comparison of Qwen3-VL-8B-Instruct~\cite{qwen3} trained with different data source under Chance-Adjusted Accuracy (CAA, \%) on the SITE-Bench~\cite{SITE}. The text in \textcolor{red}{red} and \textcolor{blue}{blue} denotes the best and second-best results.}
\label{tab:caa_comparison}
\resizebox{\textwidth}{!}{
\begin{tabular}{lccccccc}
\toprule
\textbf{Model} & \textbf{Overall} & \textbf{Cnt./Exist.} & \textbf{Obj. Loc.} & \textbf{3D Info.} & \textbf{MV/XImg.} & \textbf{Spatial Rel.} & \textbf{Move/Nav.} \\
\midrule
Baseline        & 43.54 & 55.99    & 49.11 & 56.00 & 23.05 & 66.13 & \textcolor{blue}{21.32} \\
SenseNova-SI-150K~\cite{SenseNova} & \textcolor{blue}{45.35} & \textcolor{blue}{57.36} & \textcolor{blue}{51.91} & \textcolor{red}{\textbf{58.73}} & \textcolor{blue}{24.75} & \textcolor{blue}{68.45} & 19.51 \\
\midrule
\rowcolor{pink!20}Ours-50K   & 43.94 & 56.97 & 47.07 & 56.78 &23.52 & 66.79 & \textcolor{red}{\textbf{23.74}} \\
\rowcolor{pink!20}Ours-50K+SenseNova-SI-100K~\cite{SenseNova}   & \textcolor{red}{\textbf{45.51}} &\textcolor{red}{\textbf{58.74}} & \textcolor{red}{\textbf{58.85}} & \textcolor{blue}{58.34} & \textcolor{red}{\textbf{25.78}} & \textcolor{red}{\textbf{68.61}} & 18.90 \\

\bottomrule
\end{tabular}
}

\begin{flushleft}
\footnotesize
\textbf{Abbreviations:} Cnt./Exist. = counting \& existence; Obj. Loc. = object localization \& positioning; 
3D Info. = 3D information understanding; MV/XImg. = multi-view \& cross-image reasoning; 
Spatial Rel. = spatial relationship reasoning; Move/Nav. = movement prediction \& navigation.
\end{flushleft}
\end{table}

\begin{table}[!htb]
\centering
\caption{Performance comparison on 3DSRBench~\cite{3dsrbench} under different evaluation settings. Bold indicates the better result within each model group.}
\label{tab:3dsrbench_main}
\footnotesize
\begin{tabular}{l l c c c c}
\toprule
\textbf{Model} & \textbf{Method} & \textbf{Vanilla} & \textbf{Flip} & \textbf{Circ} & \textbf{Flip+Circ} \\
\midrule
\multirow{2}{*}{Qwen3-VL-8B-Instruct~\cite{qwen3}} 
& Baseline & \textbf{0.566} & 0.498 & 0.533 & 0.475 \\
& \cellcolor{pink!15}Ours-50K     & \cellcolor{pink!15}0.564 & \cellcolor{pink!15}\textbf{0.499} & \cellcolor{pink!15}\textbf{0.541} & \cellcolor{pink!15}\textbf{0.483} \\
\midrule
\multirow{2}{*}{InternVL3-8B~\cite{internvl}} 
& Baseline & 0.524 & 0.444 & 0.396 & 0.324 \\
& \cellcolor{pink!15}Ours-150K     & \cellcolor{pink!15}\textbf{0.541} & \cellcolor{pink!15}\textbf{0.466} & \cellcolor{pink!15}\textbf{0.447} & \cellcolor{pink!15}\textbf{0.384} \\
\bottomrule
\end{tabular}
\begin{flushleft}
\footnotesize
\textbf{Abbreviations:} \emph{Vanilla} denotes the original evaluation setting; 
\emph{Flip} applies image flipping; 
\emph{Circ} applies circular rotation; 
\emph{Flip+Circ} applies both transformations together.
\end{flushleft}
\end{table}

Table~\ref{tab:caa_comparison} shows that our chained supervision transfers well to the real-world benchmark with diverse spatial sub-tasks. Although our model is trained only on simulator-generated data, it remains competitive with SenseNova-SI~\cite{SenseNova}, which is trained on substantially more real-data supervision. More importantly, our method shows a clear advantage on \emph{Move/Nav.}, where \textbf{Ours-50K} achieves the best score. This is particularly consistent with our design, since chained supervision explicitly encourages preserving and updating spatial state across dependent steps. Moreover, combining our data with SenseNova-SI-100K further improves the overall score and achieves the best results on several SITE sub-categories, suggesting that \textbf{our supervision can be complementary to existing real-data pipelines}. Table~\ref{tab:3dsrbench_main} further supports the robustness of our learned spatial reasoning under geometric transformations. For Qwen3-VL-8B-Instruct~\cite{qwen3}, our method improves three out of four evaluation settings, especially under \emph{Circ} and \emph{Flip+Circ}. For InternVL3-8B~\cite{internvl}, the gains are consistent across all four settings. Overall, these results provide additional evidence that the proposed chained design principle can transfer beyond the simulator-generated training distribution and improve several aspects of spatial reasoning.

\paragraph{Task-Level Analysis on External Spatial Benchmarks.}
To better understand where the improvement comes from, we further analyze the task-level performance on four external spatial reasoning benchmarks. As shown in Table~\ref{tab:external_task_breakdown}, the improvements are substantially larger on several higher-order spatial tasks. On VSI-Bench~\cite{vsi-bench}, ChainSpace-Pipeline improves 7 of 8 task categories, with the largest gain on route planning ($+3.56$), followed by absolute distance ($+1.54$), relative distance ($+1.30$), and appearance order ($+1.18$). On MMSI-Bench, the largest improvements occur in camera--camera relation ($+6.41$), attribute measurement ($+6.25$), appearance reasoning ($+2.98$), and multi-step spatial reasoning ($+2.02$). Similar trends are observed on ViewSpatial-Bench~\cite{viewspatialbench} and 3DSRBench~\cite{3dsrbench}, where ChainSpace-Pipeline particularly improves perspective-conditioned object orientation, scene simulation, and viewpoint-to-object relations. Overall, the gains are concentrated on tasks that require relational composition, reference-frame transformation, multi-step reasoning, or planning, which closely aligns with the capabilities targeted by chained spatial supervision.

\begin{table*}[t]
\centering
\caption{
Task-level analysis on external spatial reasoning benchmarks. ChainSpace-Pipeline-8B yields larger gains on compositional, viewpoint-conditioned, and planning-related spatial tasks.
}
\label{tab:external_task_breakdown}

\scriptsize
\setlength{\tabcolsep}{4pt}
\renewcommand{\arraystretch}{1.08}

\begin{tabularx}{\textwidth}{
    l
    >{\centering\arraybackslash}p{3.0cm}
    >{\centering\arraybackslash}p{2.8cm}
    X
}
\toprule

\textbf{Benchmark} &
\shortstack{\textbf{Qwen3-VL-8B-Instruct}\\\cite{qwen3}} &
\shortstack{\textbf{ChainSpace-Pipeline-8B}\\\textbf{(Ours)}} &
\textbf{Representative Task-Level Gains} \\

\midrule

VSI-Bench~\cite{vsi-bench}
& 57.90
& \textbf{58.82}
& Route Planning $+3.56$; 

Absolute Distance $+1.54$; 

Relative Distance $+1.30$; 

Appearance Order $+1.18$ \\

MMSI-Bench~\cite{MMSI-Bench}
& 31.10
& \textbf{31.80}
& Camera--Camera Relation $+6.41$; 

Attribute Measurement $+6.25$; 

Appearance $+2.98$; 

Multi-step Spatial Reasoning $+2.02$ \\

ViewSpatial-Bench~\cite{viewspatialbench}
& 42.20
& \textbf{42.96}
& Person-perspective Object Orientation $+3.30$; 

Scene Simulation $+0.85$ \\

3DSRBench (Circ.)~\cite{3dsrbench}
& 53.30
& \textbf{54.10}
& Viewpoint-to-Object Relation $+2.64$; 

Next-to Relation $+1.79$; 

In-front-of Relation $+1.15$ \\

\bottomrule
\end{tabularx}
\end{table*}

\paragraph{End-to-End Embodied Navigation.}
To evaluate whether the spatial capabilities learned from ChainSpace transfer beyond QA-style reasoning, we further evaluate ChainSpace-Pipeline-8B on NavBench~\cite{navbench}, an end-to-end embodied navigation benchmark requiring sequential perception, planning, and action. We evaluate 432 episodes, including 144 episodes at each of the easy, medium, and hard difficulty levels, under the same NavBench protocol without any NavBench-specific training. We additionally report the official results of GPT-4o~\cite{gpt4o} and Qwen2.5-VL-7B-Instruct~\cite{qwen25} as closed- and open-source references. As shown in Table~\ref{tab:navbench}, ChainSpace-Pipeline-8B achieves the best overall navigation performance among the evaluated open-source models, improving the execution average from 35.40 to 36.80 over the Qwen3-VL-8B-Instruct~\cite{qwen3} backbone. The improvement is particularly clear on the hard split, where the success rate increases from 26.39 to 29.86 and SPL from 20.87 to 22.82. Notably, these gains are obtained using only 50K simulator-generated chained samples, whereas SenseNova-SI-8B is trained with approximately 8M synthetic and real-world spatial samples. These results provide evidence that the spatial capability learned through chained supervision transfers to sequential embodied decision making, while also highlighting the data efficiency of organizing supervision around cross-round spatial dependencies.

\begin{table*}[t]
\centering
\caption{
End-to-end embodied navigation performance on NavBench~\cite{navbench}. All models are evaluated without NavBench-specific training. \textcolor{red}{Red} and \textcolor{blue}{blue} denote the best and second-best results among open-source models, respectively. Exec. Avg. denotes the average of SR and SPL across the three difficulty levels.
}
\label{tab:navbench}
\scriptsize
\setlength{\tabcolsep}{8pt}
\resizebox{\textwidth}{!}{
\begin{tabular}{lccccccc}
\toprule
\textbf{Model}
& \multicolumn{2}{c}{\textbf{Easy}}
& \multicolumn{2}{c}{\textbf{Medium}}
& \multicolumn{2}{c}{\textbf{Hard}}
& \textbf{Exec. Avg.} \\
\cmidrule(lr){2-3}
\cmidrule(lr){4-5}
\cmidrule(lr){6-7}
& \textbf{SR} & \textbf{SPL}
& \textbf{SR} & \textbf{SPL}
& \textbf{SR} & \textbf{SPL}
& \\
\midrule

GPT-4o~\cite{gpt4o}
& 67.36
& 54.31
& 41.67
& 35.71
& 27.78
& 21.15
& 41.33 \\

Qwen2.5-VL-7B-Instruct~\cite{qwen25}
& 41.67
& 32.55
& 22.92
& 17.43
& 10.42
& 5.67
& 21.78 \\

Qwen3-VL-8B-Instruct~\cite{qwen3}
& \textcolor{blue}{50.69}
& \textcolor{blue}{38.04}
& \textcolor{red}{\textbf{42.36}}
& \textcolor{red}{\textbf{34.05}}
& \textcolor{blue}{26.39}
& \textcolor{blue}{20.87}
& \textcolor{blue}{35.40} \\

SenseNova-SI-8B~\cite{SenseNova}
& 46.53
& 33.48
& 25.00
& 17.64
& 22.22
& 15.32
& 26.70 \\

\textbf{ChainSpace-Pipeline-8B (Ours)}
& \textcolor{red}{\textbf{53.47}}
& \textcolor{red}{\textbf{39.65}}
& \textcolor{red}{\textbf{42.36}}
& \textcolor{blue}{32.66}
& \textcolor{red}{\textbf{29.86}}
& \textcolor{red}{\textbf{22.82}}
& \textcolor{red}{\textbf{36.80}} \\

\bottomrule
\end{tabular}
}
\end{table*}

\paragraph{Robustness and interpretability of the Chain-Aware Metric.}
As shown in Table~\ref{tab:weight_ablation_full}, \textbf{(1)} \emph{our Chain-Aware Metric is robust.} The relative ranking among representative methods remains unchanged under all five weight settings. This shows that the main conclusion of our paper does not depend on a particular choice of weights.
\textbf{(2)} \emph{The metric is interpretable and flexible}. Different weight settings naturally correspond to different evaluation perspectives. For example, $W_2$ removes all chained terms and reduces the metric to per-question correctness. 
\textbf{(3)} \emph{The default setting $W_1$ is well aligned with our goal}. It provides a balanced evaluation of local correctness and chained consistency, while explicitly assigning greater importance to more complete logical chains. This matches our core motivation.

\begin{table}[!htb]
\centering
\footnotesize
\caption{
Ranking robustness under different weight settings of our \emph{Chain-Aware Metric} (Eq.~(\ref{chain-aware-metric})) on \texttt{ChainSpace-Bench}. We vary the weights of the first-order terms $(\alpha_t)$, second-order terms $(\beta_{ij})$, and the third-order term $(\gamma)$, and report the resulting normalized Avg. Chain-Aware scores for representative methods in Table~\ref{tab:chainspace_main_result}. The text in \textcolor{red}{red} and \textcolor{blue}{blue} denotes the best and second-best results.
}
\label{tab:weight_ablation_full}
\setlength{\tabcolsep}{2pt}
\renewcommand{\arraystretch}{1.15}
\begin{tabular}{lccccccc}
\toprule
\textbf{Setting} & $\boldsymbol{\alpha_1,\alpha_2,\alpha_3}$ & $\boldsymbol{\beta_{12},\beta_{23},\beta_{13}}$ & $\boldsymbol{\gamma}$ & \textbf{Qwen3-VL-8B-Instruct~\cite{qwen3}} & \textbf{RoboBrain2.5~\cite{robobrain25}} & \textbf{SenseNova-SI~\cite{SenseNova}} & \cellcolor{pink!15}\textbf{Ours} \\
\midrule
$W_1$ & (0.5,0.5,0.5) & (2.0,1.5,1.25) & 3.0 & 0.1522 & 0.1339 & \textcolor{blue}{0.1555} & \cellcolor{pink!15}\textbf{\textcolor{red}{0.1687}} \\
$W_2$ & (1.0,1.0,1.0) & (0.0,0.0,0.0) & 0.0 & 0.3343 & 0.3384 & \textcolor{blue}{0.3418} & \cellcolor{pink!15}\textbf{\textcolor{red}{0.3510}} \\
$W_3$ & (0.0,0.0,0.0) & (1.0,1.0,1.0) & 2.0 & 0.1161 & 0.0925 & \textcolor{blue}{0.1189} & \cellcolor{pink!15}\textbf{\textcolor{red}{0.1327}} \\
$W_4$ & (1.0,1.0,1.0) & (0.0,0.0,0.0) & 2.0 & 0.2298 & 0.2221 & \textcolor{blue}{0.2357} & \cellcolor{pink!15}\textbf{\textcolor{red}{0.2459}} \\
$W_5$ & (1.0,1.0,1.0) & (1.0,1.0,1.0) & 1.0 & 0.1979 & 0.1847 & \textcolor{blue}{0.2025} & \cellcolor{pink!15}\textbf{\textcolor{red}{0.2145}} \\
\bottomrule
\end{tabular}
\vspace{-0.2cm}
\end{table}


\begin{table}[!htb]
\centering
\footnotesize
\setlength{\tabcolsep}{8pt}
\renewcommand{\arraystretch}{1.1}
\caption{
Scaling behavior of chained supervision on \texttt{ChainSpace-Bench}.
We report the paired improvement $\Delta$ over the same untuned baseline under different training data scales.
Each entry is shown as $\Delta$ followed by the paired-bootstrap $p$-value in parentheses.
$^{*}$ indicates statistical significance ($p<0.05$ and 95\% CI excluding 0). The text in \textcolor{red}{red} and \textcolor{blue}{blue} denotes the best and second-best results.
}
\label{tab:scaling_curve_chainspace}
\begin{tabular}{lcccc}
\toprule
\textbf{Metric} & \textbf{25K} & \cellcolor{pink!15}\textbf{50K} & \textbf{75K} & \textbf{125K} \\
\midrule
Full-chain Exact Match 
& +0.0038 (0.3900) 
& \cellcolor{pink!15}\textcolor{red}{\textbf{+0.0153}}$^{*}$ (0.0044) 
& \textcolor{blue}{+0.0130}$^{*}$ (0.0220) 
& \textcolor{red}{\textbf{+0.0153}}$^{*}$ (0.0153) \\

Avg. Chain-Aware Score 
& +0.0038 (0.2740) 
& \cellcolor{pink!15}\textcolor{red}{\textbf{+0.0165}}$^{*}$ (<0.0001) 
& \textcolor{blue}{+0.0141}$^{*}$ (0.0032) 
& \textcolor{blue}{+0.0141}$^{*}$ (0.0048) \\
\bottomrule
\end{tabular}

\end{table}

\begin{figure}[!htb]
    \centering
    \includegraphics[width=0.8\textwidth]{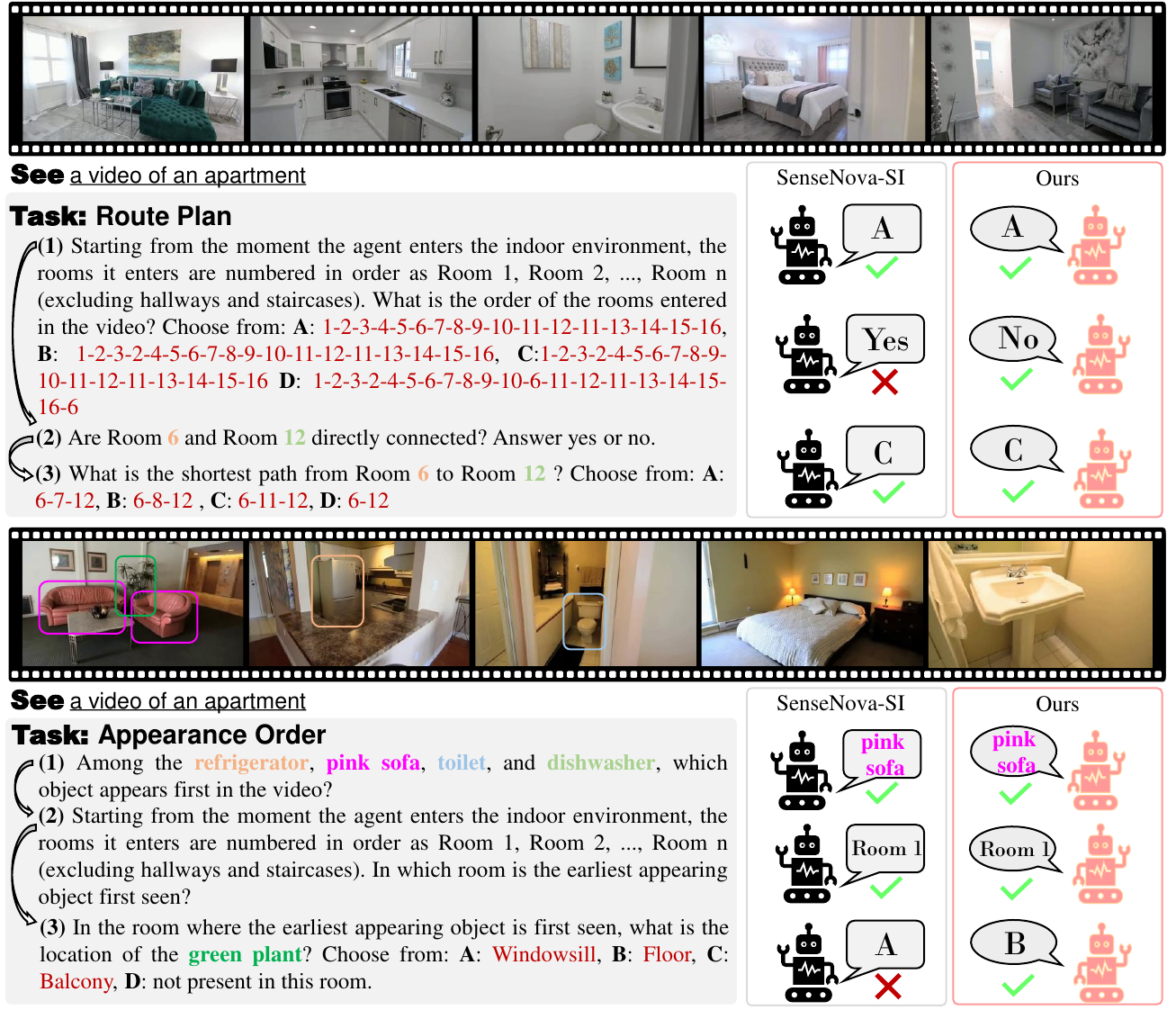}
    \caption{
    Visualization of qualitative examples on \texttt{ChainSpace-Bench}. SenseNova-SI~\cite{SenseNova} answers some individual turns correctly, whereas our model maintains consistent answers across all rounds.
}
    \label{fig:vis1}
\end{figure}

\paragraph{Scaling Behavior.}
As shown in Table~\ref{tab:scaling_curve_chainspace}, \textbf{(1)} \emph{the chain-level gains emerge early and become reliable from 50K samples onward.} At 25K, both \emph{Full-chain Exact Match} and \emph{Avg. Chain-Aware Score} already improve over the untuned baseline, and from 50K onward both metrics become consistently significant, showing that relatively small amounts of chained supervision are sufficient to induce stable improvement. 
\textbf{(2)} \emph{the early saturation is consistent with our simulator-to-real setting.} Since training is performed on synthetic trajectories while evaluation is conducted on real-world videos, additional synthetic data is naturally subject to a simulator-to-real domain gap, so gains may saturate once the core chained reasoning behavior has been learned~\cite{sim2real}.

\subsection{Qualitative Results}

\paragraph{Qualitative comparison.}
Fig.~\ref{fig:vis1} shows representative examples on \texttt{ChainSpace-Bench}. SenseNova-SI~\cite{SenseNova} answers some individual turns correctly, such as the room sequence or earliest object, but fails on intermediate or later questions that require using previously established spatial constraints. In contrast, our model answers all rounds correctly, suggesting more coherent use of accumulated spatial state.

\begin{figure}[!htb]
    \centering
    \includegraphics[width=0.8\textwidth]{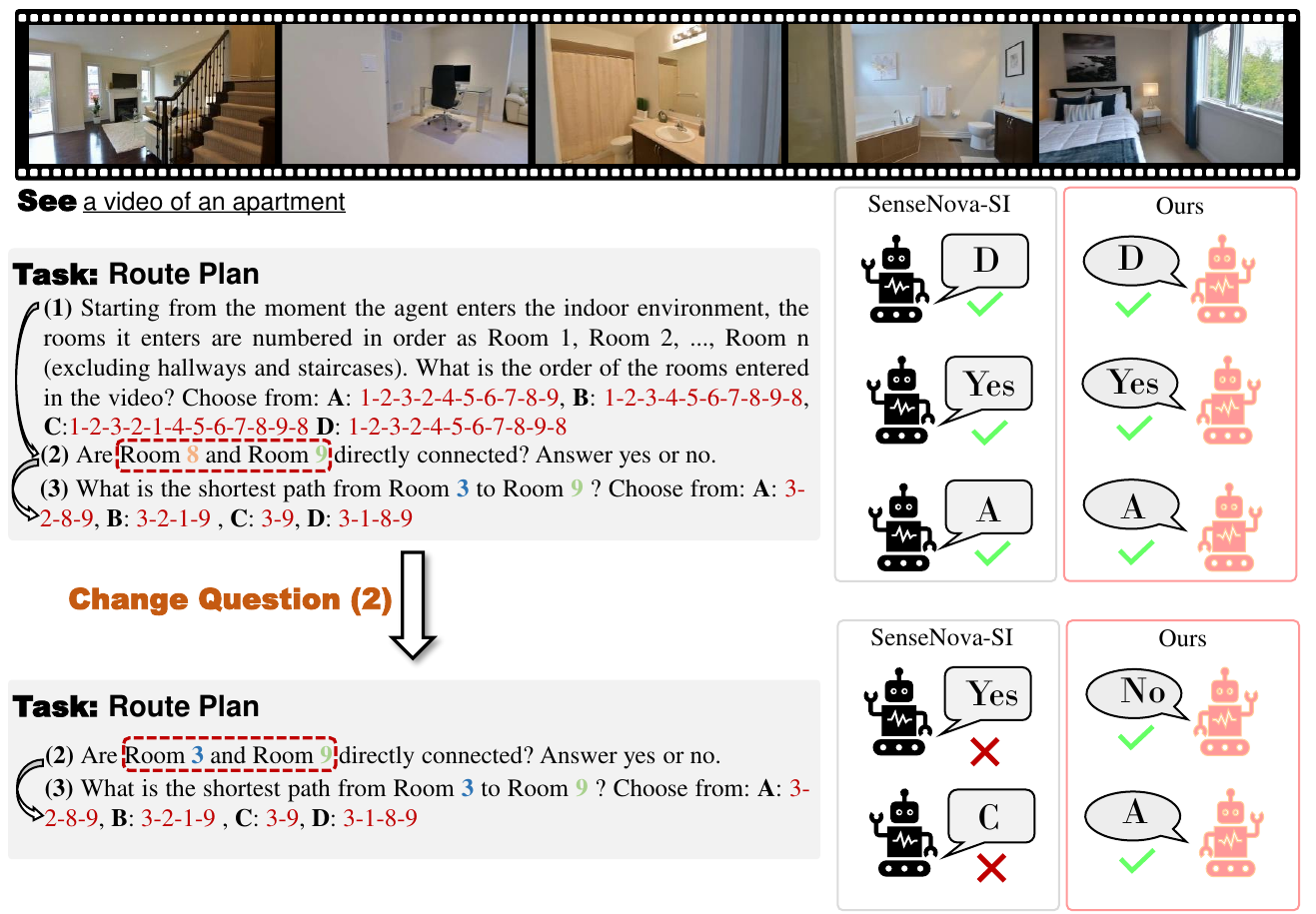}
    \caption{
    Visualization of counterfactual query examples.
    With the same video context but a changed room-pair question, our model remains correct while SenseNova-SI~\cite{SenseNova} fails, suggesting a more complete maintained spatial state.
    }
    \label{fig:vis2}
\end{figure}

\paragraph{Robustness to counterfactual spatial queries.}
Fig.~\ref{fig:vis2} tests whether the model maintains a spatial state beyond the originally queried relation. While both models answer the original route-planning chain correctly, changing the queried room pair in Question~(2) causes SenseNova-SI~\cite{SenseNova} to fail. Our model remains correct under the changed query, suggesting that \texttt{ChainSpace-Pipeline} encourages a more complete and query-flexible spatial representation.

\subsection{Failure Analysis}

\begin{figure}[!htb]
    \centering
    \includegraphics[width=0.9\textwidth]{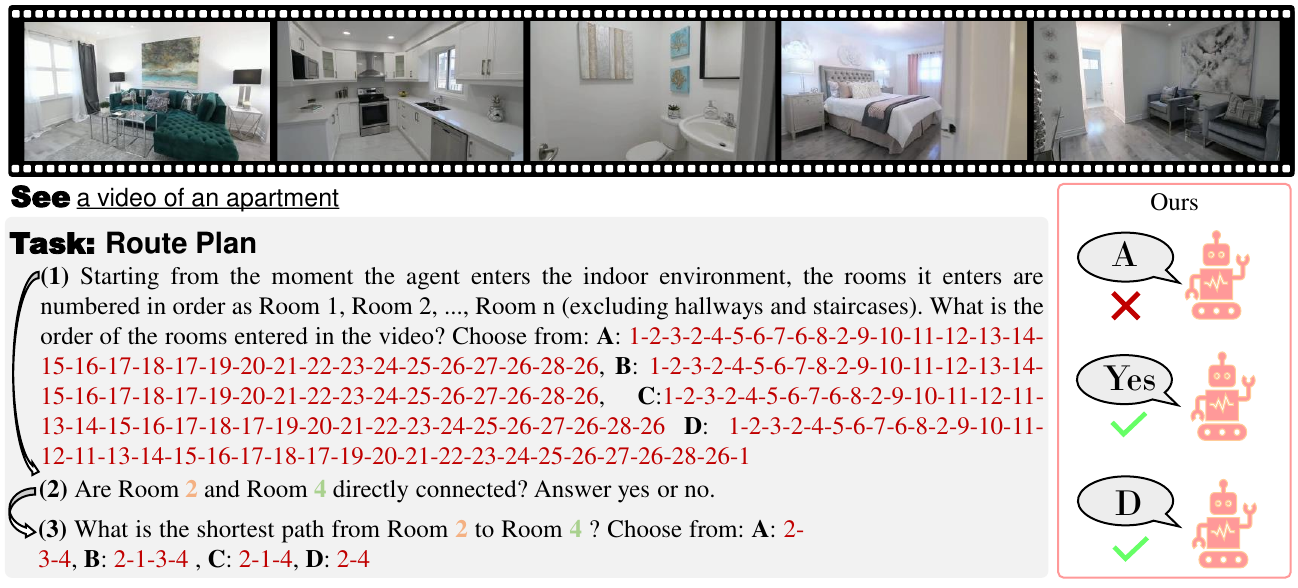}
    \caption{
    Visualization of a representative failure case. The model confuses the complete room order in a many-room trajectory, while preserving correct local spatial relations for later reasoning.
    }
    \label{fig:failure}
\end{figure}

Fig.~\ref{fig:failure} shows a long-horizon failure case that still supports the effectiveness of our method. The model fails to recover the full room visitation order in Question~(1), likely because long trajectories place a heavy burden on the long-context memory of large models~\cite{longvideobench}. However, it still correctly maintains the spatial states of the earlier queried rooms (Room~2 and Room~4), which enables correct answers to the subsequent connectivity and shortest-path questions. This suggests that \texttt{ChainSpace-Pipeline} is effective in encouraging the model to preserve spatial states that are relevant for later reasoning, even when the backbone struggles with complete global route memory.

Besides, our failure analysis reveals two observations. \emph{(1) The largest residual bottleneck lies in initial spatial-state acquisition.} Among test samples, about 67.79\% first fail at Round 1, indicating that visual grounding, temporal aggregation, and global trajectory understanding remain the main limitations of the base model. Importantly, as shown in Table 2 of the paper, ChainSpace-Pipeline significantly improves Turn-1 accuracy, showing that chained supervision strengthens initial-state establishment, although this fundamental capability still determines the current performance ceiling.

\emph{(2) Many early errors do not persist through the chain.} Among samples with an incorrect Round-1 answer, 48.98\% recover at least one later round, including 15.38\% that answer both subsequent rounds correctly. Among chains that first fail at Round 2, 32.59\% recover at Round 3. For example, the model may miss the complete room sequence in a long trajectory while still correctly preserving the local topology required for connectivity and shortest-path reasoning.

\subsection{Discussions}
\label{sec:discussion}

\paragraph{(1) Why we do not perform ablations such as reducing the chain from 3 rounds to 2 rounds, or shuffling the question order within each chain.} The reason is that these manipulations do not constitute clean controlled-variable ablations in our setting. Our benchmark and training pipeline are built on chained logical dependency, \emph{i.e.}, later questions are defined with respect to the spatial constraints established by earlier ones. As a result, changing the number of rounds or permuting the question order does not merely remove one design factor while keeping the task unchanged. Instead, it changes the task semantics and the data distribution themselves. For example, removing one round may collapse a multi-step reasoning chain into a different and often easier task, while shuffling the order can break the intended dependency structure and even make some questions ill-posed. Therefore, such experiments would not isolate the effect of chained supervision in a fair way. For this reason, we choose to support our central claim through results in Table~\ref{tab:chainspace_main_result}, Table~\ref{tab:paired_vs_baseline_fullcompare}, and Table~\ref{tab:oracle_history}. In particular, the oracle-history intervention preserves the original chained task semantics while directly probing whether later-round predictions benefit from correct preceding spatial states.

\paragraph{(2) Why we evaluate the proposed principle only under full-parameter supervised fine-tuning (SFT), rather than further introducing reinforcement learning (RL).} The main reason is that the goal of this paper is to study whether chained supervision improves the model's \emph{base spatial intelligence}. In prior practice, improvements in base capability are typically established primarily through supervised learning, while RL is more often used as a subsequent policy refinement stage~\cite{SFT_RL}. Under this perspective, SFT provides the cleanest setting for testing whether the proposed chained design principle itself yields a stronger spatial reasoning capability. A second reason is that RL is less straightforward in our setting. If RL is applied directly to model responses, the reward must largely be designed around whether the answer sequence preserves cross-question logical consistency. However, logical consistency alone does not guarantee genuine spatial intelligence. A model may produce answers that are mutually compatible as a reasoning chain, yet still be misaligned with the actual spatial structure in the video. Therefore, we choose to first isolate and validate the chained design principle through supervised learning. We believe this provides a cleaner foundation for future work on RL, where the same principle could be combined with more carefully grounded rewards to further refine spatial reasoning policies.

\paragraph{(3) Why multi-round QA instead of asking all questions at once.}
We use a multi-round QA format because our goal is to evaluate spatial reasoning as a process of progressively establishing and reusing spatial states, rather than as a static joint-QA problem. If all questions are presented simultaneously, the model can see the entire question set before answering, including future queries that may reveal which objects, rooms, or relations are important. This may allow the model to answer each question independently or use later questions as hints to narrow down earlier reasoning, without explicitly preserving an intermediate spatial state. In contrast, the multi-round format exposes the dependency structure more directly, \emph{i.e.}, each later question is conditioned on the spatial constraints established in previous turns, making it possible to evaluate whether the model can maintain and reuse them across the chain. This design also enables task-preserving analyses such as oracle-history intervention, where previous model answers can be replaced by ground-truth answers to test whether later predictions benefit from correct preceding spatial states. Therefore, the multi-round design is not a superficial prompting choice, but an essential part of our benchmark for evaluating persistent spatial reasoning.

\paragraph{(4) Discussion on the simulator-based data pipelines and Limitations.}
The main contribution of this work is the \emph{chained design principle} rather than simulation itself. The results in the manuscript already show that this principle is effective even in the simulator-based setting. However, the simulated training data inevitably introduces a simulator-to-real domain gap when evaluating on real-world videos and existing spatial intelligence benchmarks~\cite{sim2real}. Therefore, the gains on benchmarks should not be interpreted as the upper bound of chained supervision, but rather as evidence that the proposed principle can transfer beyond the synthetic training distribution. We expect larger and more consistent improvements when the chained construction principle can be applied to large-scale real-world trajectories, where the visual distribution, object arrangements, and navigation patterns are closer to practical deployment scenarios.

Meanwhile, the results in Table~\ref{tab:caa_comparison} also suggest that our simulator-generated chained data can serve as an effective complement to existing real-data pipelines. Even though the training data are fully synthetic, when combined with existing real-data supervision, they further improve performance. This indicates that the structured signal provided by our pipeline captures information that is not fully covered by current independent-QA real datasets, and therefore offers a useful supplementary source of supervision for spatial intelligence learning.

\section{Conclusion}

We presented \texttt{ChainSpace}, a paradigm for evaluating and learning spatial intelligence through chained reasoning. 
By organizing spatial reasoning as a logically dependent multi-round process, we provide both a more faithful benchmark for persistent spatial reasoning and a more informative supervision signal for training. Experiments validate the effectiveness of our design principle. 

\bibliographystyle{plainnat}
\bibliography{Ref}




\end{document}